\documentclass{article}

\usepackage[nonatbib, final]{nips_2018}

\usepackage[numbers,square]{natbib}

\usepackage{booktabs} %
\usepackage{todonotes}
\usepackage{placeins}
\usepackage{color}
\usepackage{algorithm}
\usepackage{algpseudocode}

\usepackage{geometry}
\usepackage{graphicx}

\usepackage{statistics-macros}

\usepackage[utf8]{inputenc} %
\usepackage[T1]{fontenc}    %
\usepackage{hyperref}       %
\usepackage{url}            %
\usepackage{amsfonts}       %
\usepackage{nicefrac}       %
\usepackage{microtype}      %
\usepackage{subfigure}
\usepackage{caption}

\title{In-silico Risk Analysis of Personalized Artificial Pancreas Controllers via Rare-event Simulation}
\author{
	Matthew O'Kelly\thanks{Equal contribution. This is preliminary work.} \\
	University of Pennsylvania \\
	\texttt{mokelly@seas.upenn.edu} \\
	\And
	Aman Sinha$^*$ \\
	Stanford University \\
	\texttt{amans@stanford.edu} \\
	\And
	Justin Norden$^*$ \\
	Stanford University \\
	\texttt{jnorden@stanford.edu} 
	\And
	Hongseok Namkoong$^*$ \\
	Stanford University \\
	\texttt{hnamk@stanford.edu} \\
}
\begin{document}
\maketitle

\vspace{-25pt}

\begin{abstract}
	\vspace{-10pt}
	Modern treatments for Type 1 diabetes (T1D) use devices known as \textit{artificial pancreata} (APs), which combine an insulin pump with a continuous glucose monitor (CGM) operating in a closed-loop manner to control blood glucose levels. In practice, poor performance of APs (frequent hyper- or hypoglycemic events) is common enough at a population level that many T1D patients modify the algorithms on existing AP systems with unregulated open-source software. Anecdotally, the patients in this group have shown superior outcomes compared with standard of care, yet we do not understand how safe any AP system is since adverse outcomes are rare. In this paper, we construct generative models of individual patients' physiological characteristics and eating behaviors. We then couple these models with a T1D simulator approved for pre-clinical trials by the FDA. Given the ability to simulate patient outcomes in-silico, we utilize techniques from rare-event simulation theory in order to efficiently quantify the performance of a device with respect to a particular patient. We show a 72,000$\times$ speedup in simulation speed over real-time and up to 2-10 times increase in the frequency which we are able to sample adverse conditions relative to standard Monte Carlo sampling. In practice our toolchain enables estimates of the likelihood of hypoglycemic events with approximately an order of magnitude fewer simulations.
\end{abstract}
\vspace{-20pt}

\section{Introduction}
\vspace{-10pt}
T1D impacts more than 1,000,000 people in the US
\cite{menke_prevalence_2013}. Patients are often diagnosed with T1D as
children, and for the rest of their lives they are responsible for calculating and administering doses of insulin, a potent, lethal drug
\cite{subramanian_management_2000}. The consequences of poor T1D management
are severe. Patients who receive too little insulin experience hyperglycemia,
a condition which leads to permanent organ damage. On the other hand, excessive doses of
insulin cause hypoglycemia, a potentially fatal condition~\cite{nathan_intensive_2005}.
Modern treatment systems automatically monitor blood glucose levels and deliver insulin to T1D patients \cite{subramanian_management_2000}. Commonly known as \textit{artificial pancreata} (APs), these devices combine an insulin pump with a continuous glucose monitor (CGM) operating in a closed-loop manner to control blood glucose levels.
 Fundamentally, the system defined by a T1D patient controlled with an AP is underactuated. Specifically, APs can only control insulin delivery, which lowers blood glucose levels; APs \emph{do not currently} have any way to increase blood glucose levels. Typical methods to increase blood glucose levels, like eating food, are unavailable when the patient is sleeping or otherwise incapacitated. As such, AP algorithms are conservative in order to prevent hypoglycemic episodes for which the system has no direct actuation authority to escape. 
In practice, poor performance (\textit{e.g.} hyper- or hypoglycemic events occurring too frequently) is common enough at a population level that more than 1,000 patients with T1D  have modified the algorithms on existing AP systems with unregulated open-source software \cite{lee_real-world_2017}. The patients in this group have anecdotally shown superior outcomes compared with standard of care, and yet we do not accurately know how safe the systems truly are.

In this work, we aim to uncover the probability of failure for an AP under a data-driven distribution of T1D patient behavior and physiology. Specifically, we consider the scenario of overnight fasting, a dangerous time for T1D patients. Using an FDA-approved simulator for T1D patients' physiology, we investigate the last evening meal and overnight fasting period. Our data-driven distribution consists of the carbohydrate composition of the evening meal, the fasting duration, and internal physiological parameters for a T1D patient. We use adaptive importance sampling to iteratively learn an estimator that efficiently learns the (rare) probability of a hypoglycemic event. Compared to naive Monte Carlo sampling, our adaptive importance sampling method increases the frequency of sampling rare hypoglycemic events and more accurately estimates the probability of these events. Indeed, for the same number of samples, we find 2-10$\times$ as many hypoglycemic events and our estimates of the probability of these events have 2-4$\times$ smaller variance.

\section{Simulator \& Population Modeling}
\label{sec:sim}
\vspace{-10pt}

\label{sec:gen}

\paragraph*{Simulator} We use an implementation of the 2008 UVa Padova simulator~\cite{dalla_man_gim_2007} for simulating T1D patients. This simulator is composed of a system of ordinary differential equations modeling the internal dynamics of of a patient. The system is composed primarily of three subsystems: glucose physiology, insulin physiology, and carbohydrate ingestion physiology. The simulator has been approved for preclinical trials by the FDA~\cite{bergenstal_safety_2016}.
\paragraph*{Eating and Overnight Fasting Behavior}
Because we are concerned with the overnight behavior of an AP, we build a distribution of evening meals and overnight fasting times of T1D patients. Unfortunately, there are few publicly available surveys of T1D behavior. Therefore, we estimate a sample using the National Health and Nutrition Examination Survey (NHANES), a representative cross-sectional survey of the US population \cite{nhanes}. Nevertheless, even NHANES does not explicitly classify whether patients have type 1 or type 2 diabetes. As outlined by  \citet{menke2013prevalence}, we define T1D individuals in NHANES as those who started insulin within 1 year of diabetes diagnosis, are currently using insulin and were diagnosed with diabetes under age 40. 
NHANES collects two consecutive days' worth of eating behavior for survey participants, including mealtimes as well as nutritional and caloric information of each meal. 
We define meals as ingestion of food that contains at least 1 calorie, and we define overnight fasting as the longest period of fasting greater than 5 hours that starts after 4pm on the first day and ends before 1pm on the second day. We fit a logit-normal distribution to these two variables since it has the ability to approximate the shape of beta and normal distributions while also having compact support. The latter fact is important since we want to prevent unrealistic tail behavior. The details of the model are in Appendix~\ref{appendix:mealgen}.

\paragraph*{Physiological Parameters}
Thirty sample T1D patients (subpopulations of 10 children, 10 adolescents, 10 adults) are available for use with the UVa Padova simulator~\cite{dalla_man_gim_2007}. 
The rarity of data for T1D necessitates a more personalized approach to analyzing risk and certifying safety of artificial pancreata. As such, we build generative models for physiological parameters focusing on small variations in the 61 parameters per patient, since we know that each of the thirty realizations of parameters is in fact realistic. We design logit-normal distributions around each patient's parameters by first setting the compact interval range as 1/10 the interval range for the entire subpopulation centered at the patient's parameters. The details can be found in Appendix~\ref{appendix:physio}.
This covariance structure we derive encodes the fact that we have no knowledge of the covariance of the small variations in the 61 parameters for an individual patient.

\section{Rare-event simulation}
\label{sec:risk}

\vspace{-10pt}
Personalized clinical evaluation of an AP for a particular
patient profile---especially one modified from its original settings---is
infeasible due to the high costs and inherent dangers posed to trial subjects. Formal verification methods are challenging to apply due to the difficultly in specifying the operating domain. For example, a patient may have run a marathon the previous day,
dramatically increasing her insulin sensitivity. 

Instead of introducing completely ad-hoc restrictions to the set of scenarios under consideration, we consider a data-driven a probabilistic approach. We posit a base distribution $X \sim P_0$ on the set of initial conditions $\mc{X}$, and
denote by $f: \mc{X} \to \R$ a continuous measure of risk; a natural measure
for an artificial pancreas is given by the minimum blood glucose level over a
period of interest. Then, our goal is to evaluate the probability of an
adverse event
\begin{equation*}
  p_\gamma \defeq \P_0(f(X) \le \gamma),
\end{equation*}
based on samples $X_1, X_2, \ldots$ and rollouts $f(X_1), f(X_2), \ldots$ from
our simulator. Here, the parameter $\gamma > 0$ is a threshold that defines an
``adverse event''; if $f(x)$ denotes the minimum glucose level over a rollout,
then $\gamma = 70$ is the threshold for a clinical definition of
a hypoglycemic event \cite{cryer2010hypoglycemia}.

Our probabilistic approach rests on the notion of a base distribution
$P_0$, which could introduce subjective valuations of what
constitutes a realistic scenario. To alleviate these issues in a systematic
manner, we use data-driven model whenever population level distributions are
deemed to be meaningful at the individual level as well (e.g. amount of
carbohydrates consumed per meal). To model variations in personalized
measurements, we account for sensor noise and uncertainty in physiological parameters by postulating a logit-normal
distribution with small amounts of variance. We combine: (1) data-driven
estimation approach for core exogenous sources of uncertainties, \textit{e.g.} meals and fasting (2) autoregressive moving average process noise for sensor measurements (3) white noise to estimates of unobservable physiological parameters. Through this effort we are able to represent three different
sources of randomness in our base distribution $P_0$.

For reliable control algorithms, an adverse event will be rare, and the
probability $p_{\gamma}$ close to $0$. We treat this as a rare event
simulation problem (see~\cite[Chapter VI]{AsmussenGl07} for an
overview of this topic), and use adaptive importance sampling techniques to
accelerate our evaluation. To address the shortcomings of the naive Monte Carlo method for estimating
rare event probabilities $p_{\thresh}$, we use an adaptive importance sampling
approach~\cite{AsmussenGl07}. Our goal is to find an importance sampling
distribution that produces estimates with low variance. The optimal
importance-sampling distribution for estimating $p_{\thresh}$ is given by the
conditional density
$p\opt(x) = \indic{\obj(x) \le \thresh} p_0(x) / p_{\thresh}$, where $p_0$ is
the density function of $P_0$. Indeed, since $p_0(x) / p\opt(x) = p_{\thresh}$
if $\obj(x) \le \thresh$, the estimate
$\what{p}_{N, \thresh}\opt \defeq \frac{1}{N} \sum_{i=1}^N
\frac{p_0(X_i)}{p\opt(X_i)} \indic{\obj(X_i) \le \thresh}$ is exact.  However,
sampling from this distribution requires knowledge of $p_{\thresh}$, the
quantity under estimation. Instead, we consider a family of parameterized
importance sampling distributions $P_{\theta}$ for $\theta \in \Theta$, and
use a model-based optimization method that iteratively modifies $P_\theta$ to
better approximation $P\opt$.

In particular, we use the cross-entropy method~\cite{RubinsteinKr04}, which
iteratively approximates
$\theta\opt \in \argmin_{\theta \in \Theta} \dkl{P\opt}{P_{\theta}}$, the
projection of $P\opt$ onto the class of parameterized distributions
$\mc{P} = \{ P_{\theta} \}_{\theta \in \Theta}$. We use natural exponential
families as our model class $\mc{P}$ of importance samplers.

Since adverse events $\{ \obj(X) \le \thresh\}$ are rare, the cross-entropy
method maintains a surrogate distribution
$q_k(x) \propto \indic{\obj(x) \le \gamma_k} p_0(x)$ where
$\gamma_k \ge \gamma$ is a (potentially random) sequence of alternative
thresholds $\gamma$. Using this multi-level approach, at each iteration we use
samples from our current iterate $P_{\theta}$ to update $\theta$ as an
approximate projection of $Q_k$ onto $\mc{P}$. Such a procedure is guided
towards distributions $P_{\theta}$ that upweights regions of $\mc{X}$ with low
values of $\obj(x)$ (unsafe regions). To choose the level $\gamma_k$ at each
iteration $k$, we use an empirical estimate of the $\tol$-quantile of $f(X)$
where $X \sim P_{\theta_k}$, where $\tol \in (0, 1)$
(see~\cite{HomemDeMello07} for other variants). Further details are provided in Appendix~\ref{appendix:cex}.

The cross-entropy method (Algorithm~\ref{alg:ce} in Appendix \ref{appendix:cex}) is a model-based heuristic
for approximating the optimal importance sampler,
$\theta\opt \in \argmin_{\theta \in \Theta} \dkl{P\opt}{P_{\theta}}$. Empirically, we find that with careful choice of hyperparameters, the
cross-entropy method learns importance samplers that samples risky scenarios
much more frequently. We observe significant improvements over the
naive Monte Carlo method in our experiments, but we observe that the importance sampler focuses on a particular
``risk mode'' of over-eating.

\begin{figure*}[!t]

  \includegraphics[width=.32\textwidth]{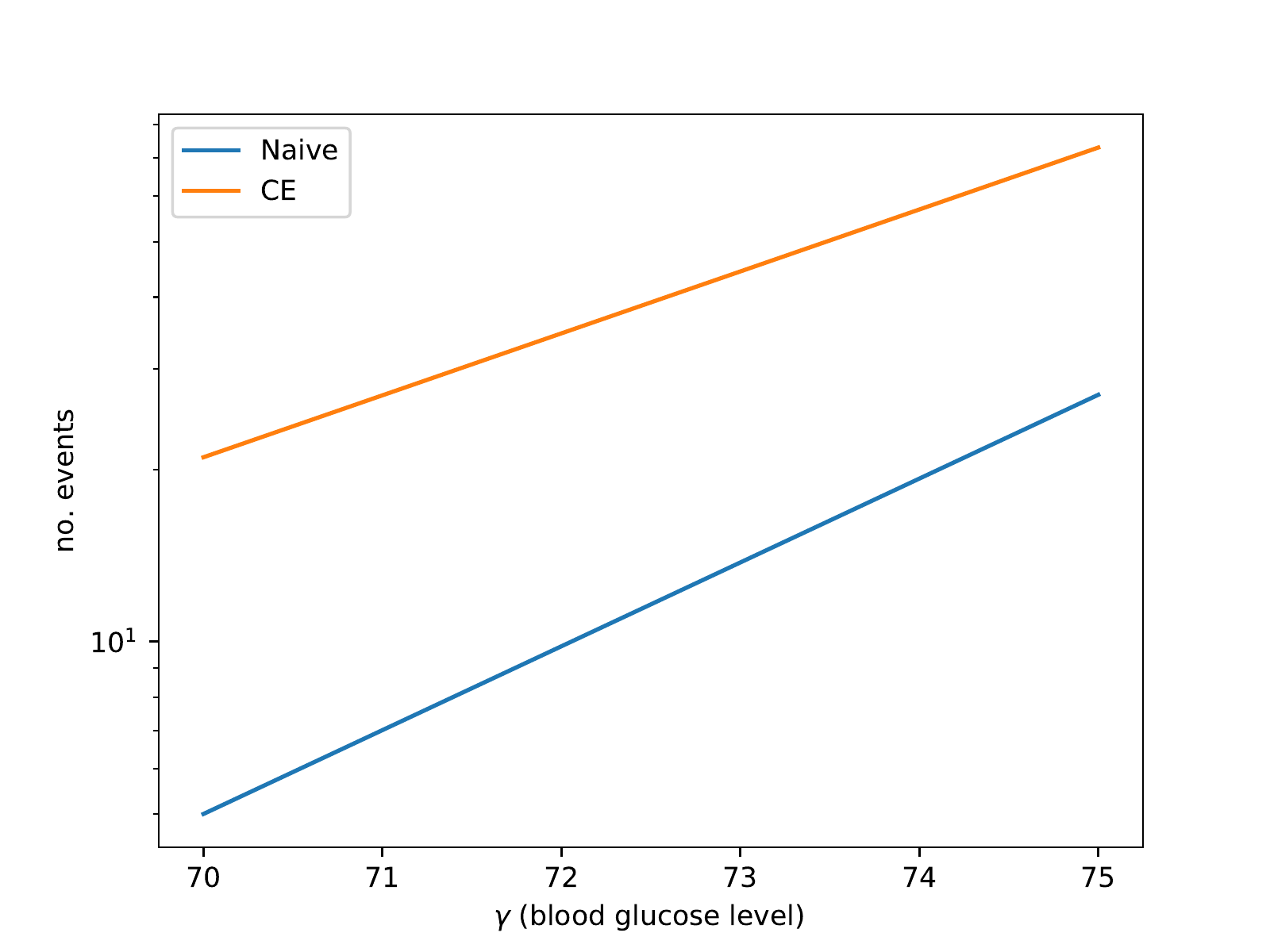}
  \includegraphics[width=.32\textwidth]{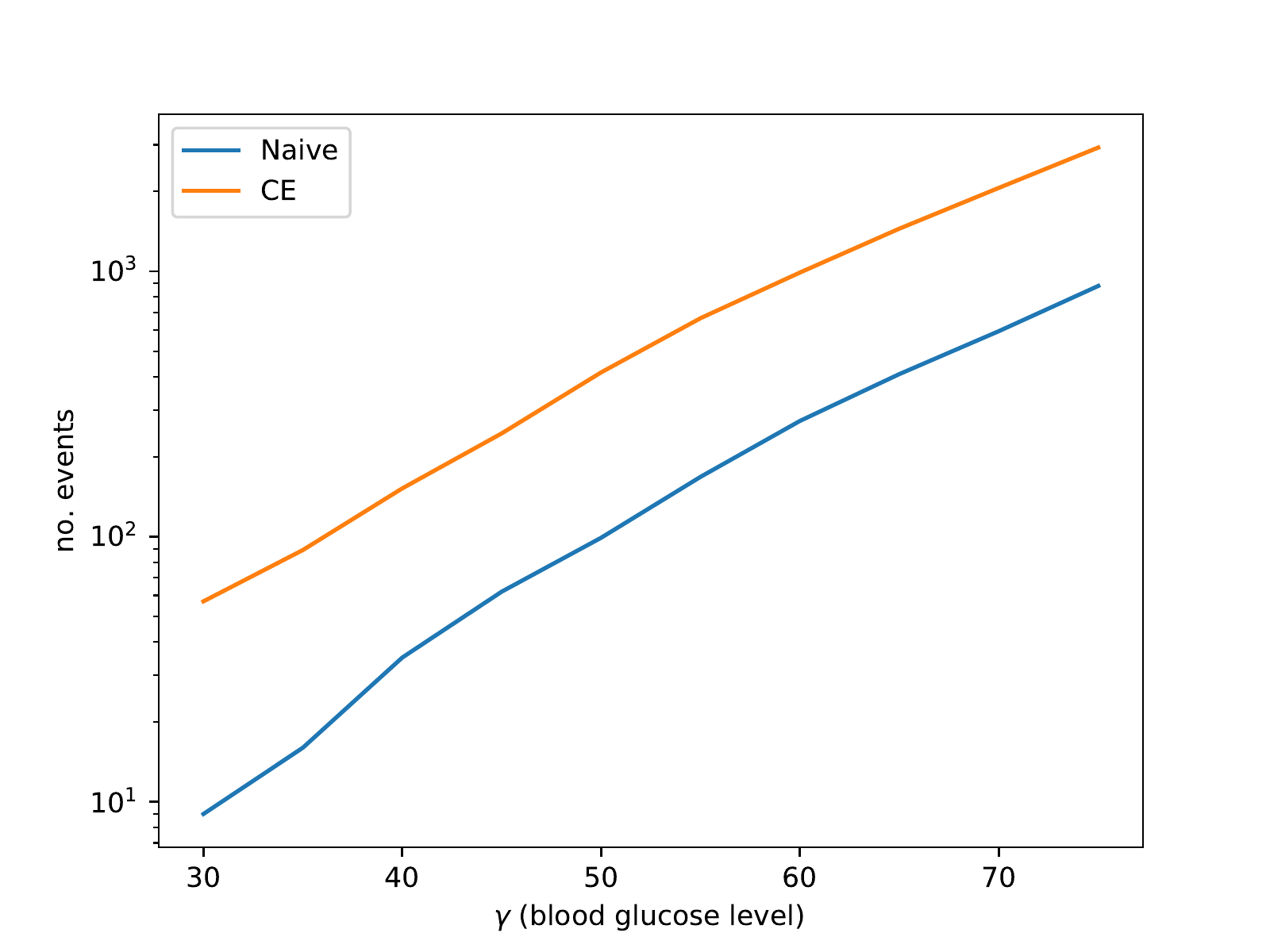}
  \includegraphics[width=.32\textwidth]{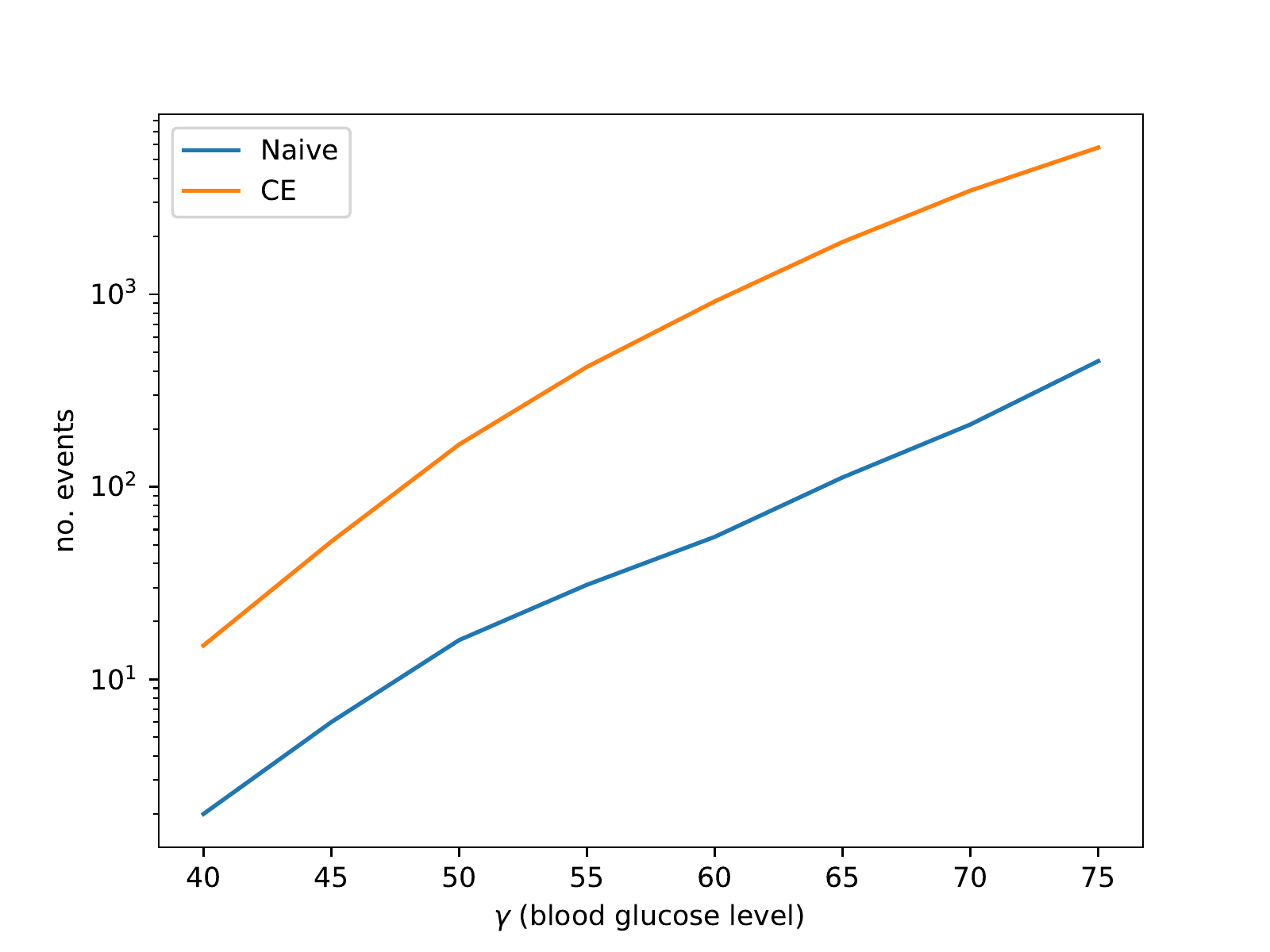}
  \caption{Number of hypoglycemic events $f(X) \le \gamma$ out of $100K$
    samples.  Cross-entropy method (orange) learns to sample $2$-$10$ times
    more adverse events than naive Monte Carlo (blue). }
  \label{fig:events}
\end{figure*}

\begin{figure*}[!t]
  
  \includegraphics[width=.32\textwidth]{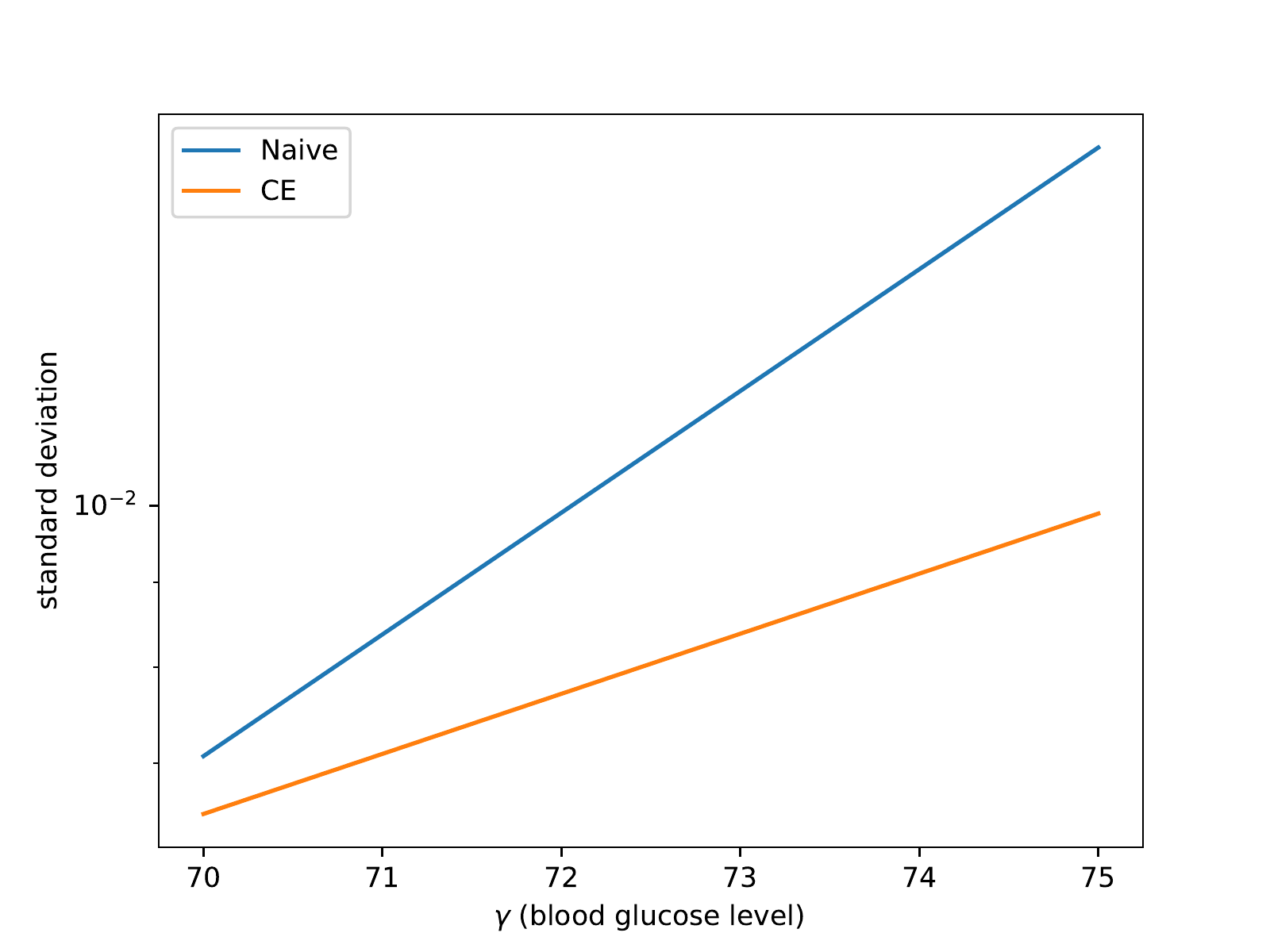}
  \includegraphics[width=.32\textwidth]{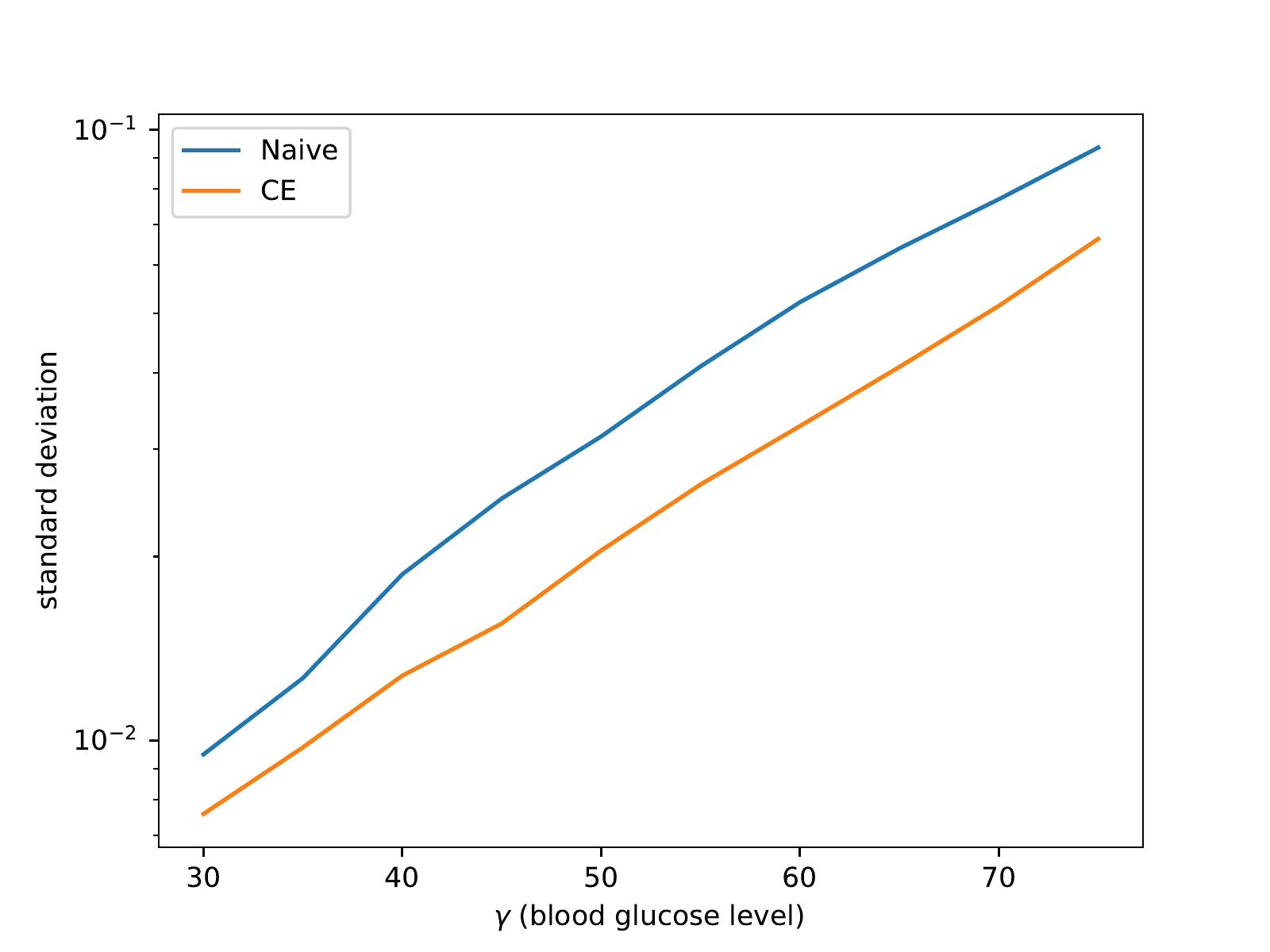}
  \includegraphics[width=.32\textwidth]{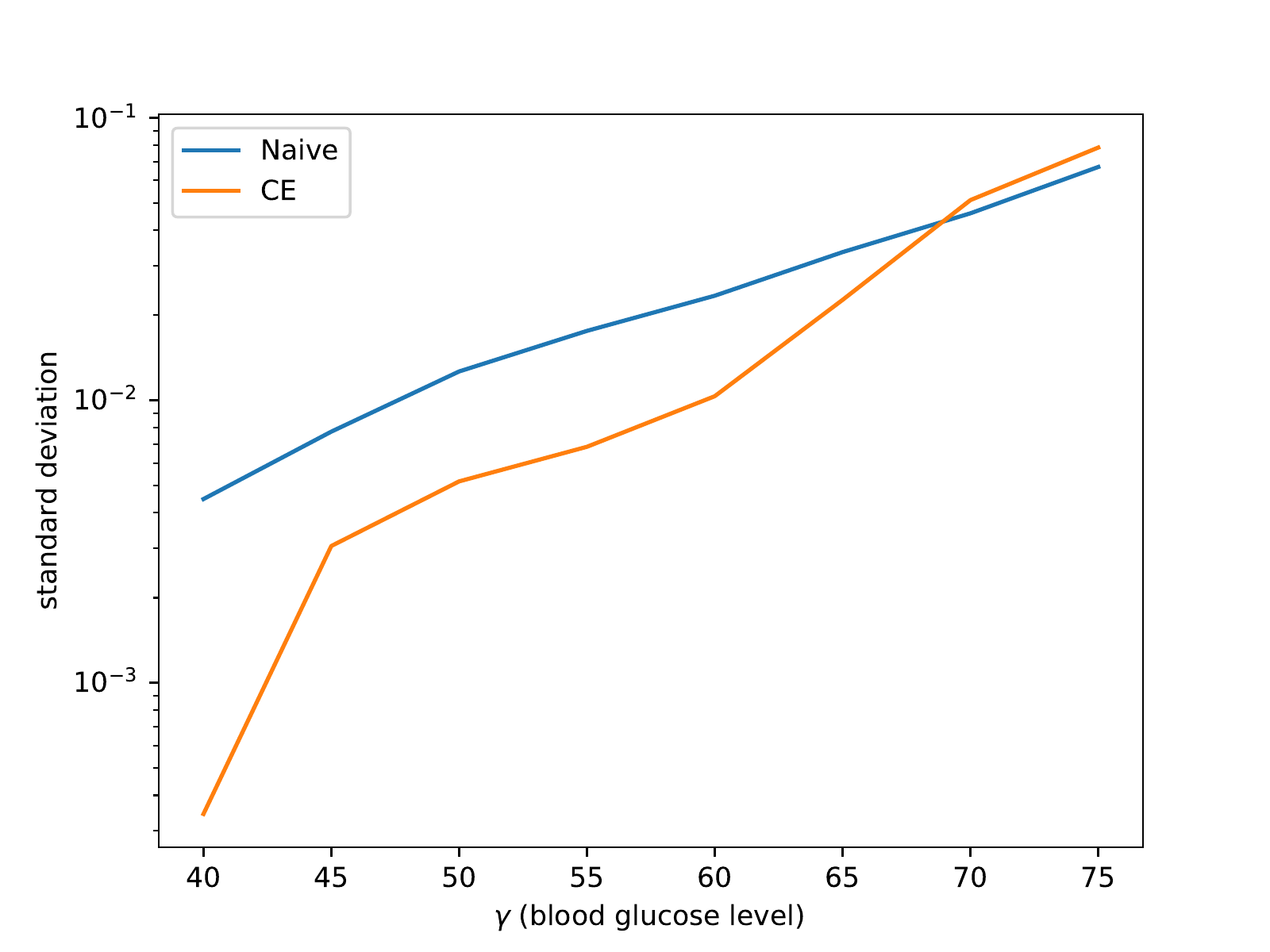}
  \caption{Standard deviation of empirical estimators for
    $p_{\gamma} = P_0(f(X) \le \gamma)$ based on $100K$ samples. By virtue of
    sampling more adverse-events, the importance sampling estimator based on
    the cross-entropy method (orange) achieves $2$-$4$ times variance reduction
    over naive Monte Carlo (blue). }
\label{fig:std}
\end{figure*}

\section{Results and Discussion}
\label{sec:experiment}
\vspace{-10pt}
We investigate hypoglycemic events as failure modes for AP systems. Recall that APs are underactuated systems which currently have the ability to lower, but not raise, blood glucose levels. As such, we choose to simulate an evening meal followed by an overnight fast to capture the one of the most dangerous time periods for a T1D patient, a period when they are incapacitated and cannot react to hypoglycemia \cite{hovorka_overnight_2011}. There are three classes of patients: children (under age 13), adolescents (between ages 13 and 19), and adults (above age 19). We show in the sequel, the frequency of hypoglycemic events subsequently varies considerably between children and both adolescents and adults. 

We let the random variable $X \in R^{63}$ be composed of $X_{up}$ concatenated with $X_{fc}$, such that a specific realization of $X$ defines a synthetic patient ($X_{up}$) as well as their evening meal carbohydrate intake and overnight fasting time ($X_{fc}$).
We define the risk $f(X)$ as the minimum blood glucose level for a patient during their simulated overnight fast. We choose a subset of one child, one adolescent, and one adult form our population of 30 patients. For each patient, we compute both a naive Monte Carlo estimate and an estimate utilizing the framework described in Section \ref{sec:risk}.

Since the feature vector
$X \sim P_0$ is a logit-normal distribution, we posit the space of
logit-normal distributions as the model space searched over by the
cross-entropy method.
Concretely, we consider logit-normal distributions with the same covariance
structure. Formally, this is equivalent to taking a logistic transform of $X$
and searching over multivariate Gaussian distributions $P_{\theta}$ with covariance
$\what{\Sigma}$ estimated in Section~\ref{sec:gen}. Specifically, we consider the search space $\Theta_r = \{ \theta: \ltwo{\theta - \what{\mu}} \le r \}$ where $\what{\mu}$ is the mean vector estimated in Section~\ref{sec:gen}, and we
choose $r$ to maximize acceleration while retaining numerical stability of the likelihood ratios.

We use $\tol = 0.01$, $\alpha_k =.8$ and $N_k = 1000$ samples per iteration of
the cross-entropy method. We observe that $\rho$ is a crucial hyperparameter
for achieving acceleration; we chose $\rho = 0.01$ based on the number of
adverse events sampled for the adult patient, and used it throughout all age
groups.  We note that in order to avoid overfitting to the $N_k$ samples drawn
at each cross-entropy iteration, one has to increase $N_k$ appropriately as
$\tol$  approaches 0. Another crucial design choice that determine the
performance of the cross-entropy method is $r$, size of the search space.
Based on a small preliminary experiment, we fix them at $\{.1, .1, .5\}$ for
child, adolescent, and adult patients.

To evaluate the performance of our learned importance sampling distribution
$P_{\what{\theta}}$, we draw $n = 100,000$ samples
$X_i \simiid P_{\what{\theta}}$ to simulate the minimum glucose level.  In
Figure~\ref{fig:events}, we find that the cross-entropy method learns to
sample hypoglycemic events $2$-$10$ times more frequently compared to the
naive Monte Carlo method. We see in Figure~\ref{fig:std} that our ability to
sample more adverse events leads to estimators with smaller variance. Since
both naive Monte Carlo and importance sampling estimators are unbiased, the
observed variance reduction implies that the cross-entropy method can
significantly accelerate evaluation of adverse events. 
The relative variance reduction is
especially pronounced for the patient in the child group, whose probability of
the adverse event is rarest out of the three patients.

 \newpage
\bibliographystyle{abbrvnat}
\bibliography{glucagon_bib}

\appendix

\appendix
\FloatBarrier
\section{Generative Models}
\subsection{Eating and Overnight Fasting Behavior}
\label{appendix:mealgen}
We fit a distribution to the data in the following manner. Denoting $Y_{fc} \in \R^2$ as the random variable denoting overnight fasting time and the carbohydrate intake of the last meal prior to fasting, we denote random variable $X_{fc}$ with parameters $a_{fc}, b_{fc},u_{fc},$ and $\Sigma_{fc}$ as follows:
\begin{subequations}
	\begin{align}
		X_{fc} & \sim \mathcal{N}\left ( \mu_{fc}, \Sigma_{fc} \right )\\
		Y_{fc} & =  (b_{fc}-a_{fc}) \;\times\; \sigma \left (X_{fc} \right) \; +\; a_{fc}\\
		\sigma(t) &= \frac{1}{1+\exp({-t})}
	\end{align}
\end{subequations}
where $\mathcal{N}(\cdot, \cdot)$ is a multivariate normal distribution, and the operators $\sigma(\cdot)$, $\times$, and $+$ operate elementwise on vectors. The four parameters are fit from the empirical data distribution $\left \{X_{fc}^i\right \}_{i=1}^n$ with its corresponding distribution $P_{e}$ that places weight $1/n$ on each datapoint.
\begin{subequations}\label{eq:logitnormal}
	\begin{align}\label{eq:logita}
		\hat{a}_{fc} &= \min_{i}Y_{fc}^i\\\label{eq:logitb}
		\hat{b}_{fc}&=\max_{i}Y_{fc}^i\\\label{eq:logitmean}
		\hat{\mu}_{fc} &= \E_{P_{e}}\sigma^{-1}\left (\frac{Y_{fc}-\hat{a}_{fc}}{\hat{b}_{fc}-\hat{a}_{fc}} \right )\\\label{eq:logitcov}
		\hat\Sigma_{fc} &= \operatorname{Cov}_{P_{e}}\left (\sigma^{-1}\left (\frac{Y_{fc}-\hat{a}_{fc}}{\hat{b}_{fc}-\hat{a}_{fc}} \right ), \sigma^{-1}\left (\frac{Y_{fc}-\hat{a}_{fc}}{\hat{b}_{fc}-\hat{a}_{fc}} \right ) \right ),\\
		\sigma^{-1}(t) &= \log\left (\frac{t}{1-t} \right ),
	\end{align}
\end{subequations}
where $\sigma^{-1}(\cdot)$, $\min(\cdot)$ and $\max(\cdot)$ operate elementwise on vectors.
\subsection{Physiological Parameters}
\label{appendix:physio}
We attempted to build a generative model of the 61 parameters $Y_{up} \in \R^{61}$ required by the simulator by fitting a logit-normal distribution to each of the three subpopulations of parameters in a manner similar to that in Equation \eqref{eq:logitnormal}. Namely, we fit $\hat a_{up}$, $\hat b_{up}$, and $\hat \mu_{up}$ using the same approach as in Equations \eqref{eq:logita}, \eqref{eq:logitb}, and \eqref{eq:logitmean}. Due to the high dimensionality of the data and the low sample size, we replace Equation \eqref{eq:logitcov} with the graphical lasso algorithm \cite{friedman2008sparse}. This method estimates the inverse covariance matrix with the following convex optimization problem \cite{friedman2008sparse}:
\begin{align*}
\hat\Sigma^{-1}_{up} &= \argmin_{T \succeq 0} \left(\tr(ST) - \log\det T + \lambda \sum_{j\neq k} |T_{jk}| \right),\\
S &= \operatorname{Cov}_{P_{e}}\left (\sigma^{-1}\left (\frac{Y_{fc}-\hat{a}_{fc}}{\hat{b}_{fc}-\hat{a}_{fc}} \right ), \sigma^{-1}\left (\frac{Y_{fc}-\hat{a}_{fc}}{\hat{b}_{fc}-\hat{a}_{fc}} \right ) \right ),
\end{align*}
However, roughly 40\% of 1,000,000 synthetic patients drawn from this model spontaneously become dangerously hypoglycemic in short periods of time without taking insulin or eating any meals, implying that the distribution is unrealistic (at least for use with the simulator).  Attempts at refining the sampling scheme using convex-hulls around the 60\% of realistic synthetic patients were not successful due to the high dimensionality of the space. Because the interdependencies between the 61 parameters are evidently paramount to ensuring realistic synthesis of patients, we move away from attempting to build accurate subpopulation distributions given only 10 sample patients for each subpopulation.

As noted in the main text, we design logit-normal distributions around each patient's parameters by first setting the compact interval range as 1/10 the interval range for the entire subpopulation centered at the patient's parameters. Namely, for patient $i$, we have:
\begin{subequations}
\begin{align}
\hat a_{up}^i &= Y^i_{up}-\frac{\hat b_{up} - \hat a_{up}}{20}\\
\hat b_{up}^i &= Y^i_{up}+\frac{\hat b_{up} - \hat a_{up}}{20},
\end{align}
where the operator $-$ is elementwise over vectors.\footnote{For some dimensions, $\hat a_{up}^i$ must be set as $ \left (Y^i_{up}-\frac{\hat b_{up} - \hat a_{up}}{20}\right )_{+}$ since the parameter must be nonnegative.}
 Then, we choose the mean and covariance as follows:
 \begin{align}
 \hat \mu_{up}^i = \sigma^{-1}\left (\frac{Y_{up}^i-\hat{a}^i_{up}}{\hat{b}^i_{up}-\hat{a}^i_{up}} \right )\\
 \hat \Sigma_{up}^i = 0.25I,
 \end{align}
\end{subequations}
This covariance structure encodes the fact that we have no knowledge of the covariance of the minute variations in the 61 parameters for an individual patient; it approximates a normal distribution with roughly 99\% of the probability mass in the interval $[\hat a_{up}^i+ 0.2(\hat{b}^i_{up}-\hat{a}^i_{up}), \hat b_{up}^i - 0.2(\hat{b}_{up}^i-\hat{a}_{up}^i)]$.

\section{The Cross-entropy method}
\label{appendix:cex}
Concretely, consider the following updates to the parameter vector $\theta_k$
at iteration $k$: compute projections of a mixture of $Q_k$ and $P_{\theta_k}$
onto $\mc{P}$
\begin{align}
	\theta_{k+1}
	& = \argmin_{\theta \in \Theta} \dkl{\alpha_k Q_k +
		(1-\alpha_k)P_{\theta_k}}{P_{\theta}} \nonumber \\
	& = \argmax_{\theta \in \Theta} \left\{ \alpha_k \E_{Q_k}[\log p_{\theta}(X)]
	+ (1-\alpha_k) \E_{\theta_k}[\log p_{\theta}(X)] \right\} \nonumber \\
	& = \argmax_{\theta \in \Theta} \left\{
	\alpha_k \theta^\top \E_{Q_k}[\suff(X)]
	+ (1-\alpha_k) \theta^\top \nabla A(\theta_k) - A(\theta) \right\}.
	\label{eqn:ideal-ce}
\end{align}
However, the $\E_{Q_k}[\suff(X)]$ term is unknown so we use an empirical
approximation. For $X_{k, 1}, \ldots, X_{k, N_k} \simiid P_{\theta_k}$, letting
$\thresh_k$ be the $\tol$-quantile of
$\obj(X_{k, 1}), \ldots, \obj(X_{k, N_k})$ and 
\begin{align}
	D_{k+1} & \defeq \frac{1}{N_k} \sum_{i=1}^{N_k}
	\frac{q_k(X_{k, i})}{p_{\theta_k}(X_{k, i})} \suff(X_{k, i}) \nonumber\\
	& = \frac{1}{N_k} \sum_{i=1}^{N_k}
	\frac{p_0(X_{k, i})}{p_{\theta_k}(X_{k, i})} \indic{\obj(X_{k, i}) \le \thresh_k}
	\suff(X_{k, i}),
	\label{eqn:is-ce}
\end{align}
we use $D_{k+1}$ in place of $\E_{Q_k}[\suff(X)]$ in the idealized
update~\eqref{eqn:ideal-ce}. Summarizing this procedure, we obtain
Algorithm~\ref{alg:ce}; as our final importance sampler, we choose $\theta_k$
with the lowest $\rho$-quantile of $f(X_{k, i})$.

\begin{algorithm}[t!]
  \caption{Cross-Entropy Method}
  \label{alg:ce}
  \begin{algorithmic}[1]
    \State Input: Quantile $\tol \in (0, 1)$,
    Stepsizes $\{\alpha_k\}_{k \in \N}$, Sample sizes $\{N_k \}_{k \in \N}$,
    Number of iterations $K$
    \State Initialize: $\theta_0 \in \Theta$
    \For{$k = 0, 1, 2, \dots, K-1$}
    \State Sample $X_{k, 1}, \ldots, X_{k, N_k} \simiid P_{\theta_k}$
    \State Set $\gamma_k$ as the minimum of $\gamma$ and the
    $\tol$-quantile of
    $f(X_{k, 1}), \ldots, f(X_{k, N_k})$
    \State
      $\theta_{k+1}
      = \argmax_{\theta \in \Theta} \left\{
    \alpha_k \theta^\top D_{k+1}
    + (1-\alpha_k) \theta^\top \nabla A(\theta_k) - A(\theta) \right\}$
    \EndFor
  \end{algorithmic}
\end{algorithm}

\end{document}